\documentclass[twocolumn, switch]{article} 

\usepackage{preprint}

\usepackage{amsmath, amsthm, amssymb, amsfonts}

\usepackage[numbers,square]{natbib}
\usepackage[utf8]{inputenc}	
\usepackage[T1]{fontenc}	
\usepackage{xcolor}		
\usepackage[colorlinks = true,
            linkcolor = purple,
            urlcolor  = blue,
            citecolor = cyan,
            anchorcolor = black]{hyperref}	
\usepackage{booktabs} 		
\usepackage{nicefrac}		
\usepackage{microtype}		
\usepackage{lineno}		
\usepackage{float}			
\usepackage{multicol}		

\usepackage{graphicx}
\usepackage{placeins}
\usepackage{subcaption} 
\usepackage[font=small,labelfont=bf]{caption} 
\usepackage{titlesec}
\titlespacing\section{0pt}{12pt plus 3pt minus 3pt}{1pt plus 1pt minus 1pt}
\titlespacing\subsection{0pt}{10pt plus 3pt minus 3pt}{1pt plus 1pt minus 1pt}
\titlespacing\subsubsection{0pt}{8pt plus 3pt minus 3pt}{1pt plus 1pt minus 1pt}

\usepackage{tikz,xcolor,hyperref}

\definecolor{lime}{HTML}{A6CE39}
\DeclareRobustCommand{\orcidicon}{
	\begin{tikzpicture}
	\draw[lime, fill=lime] (0,0) 
	circle [radius=0.16] 
	node[white] {{\fontfamily{qag}\selectfont \tiny ID}};
	\draw[white, fill=white] (-0.0625,0.095) 
	circle [radius=0.007];
	\end{tikzpicture}
	\hspace{-2mm}
}
\foreach \x in {A, ..., Z}{\expandafter\xdef\csname orcid\x\endcsname{\noexpand\href{https://orcid.org/\csname orcidauthor\x\endcsname}
			{\noexpand\orcidicon}}
}

\title{Investigating the diversity and stylization of contemporary user generated visual arts in the complexity entropy plane}

\usepackage{authblk}

\author[1,+]{Seunghwan Kim\orcidA{}}
\author[2,3,+]{Byunghwee Lee\orcidB{}}
\author[1,*]{Wonjae Lee\orcidC{}}

\affil[1]{Graduate School of Culture Technology, KAIST, Daejeon 34141, Republic of Korea.}
\affil[2]{Center for Complex Networks and Systems Research, Luddy School of Informatics, Computing, and Engineering, Indiana University, Bloomington, IN 47408, USA.
}
\affil[3]{Institute of Basic Science, Sungkyunkwan University, Suwon 16419, Republic of Korea.}
\affil[+]{S.K. and B.L. contributed equally to this work.}
\affil[*]{Corresponding author, E-mail address: wnjlee@kaist.ac.kr}

\begin{document}

\twocolumn[ 
  
\maketitle

\begin{abstract}
The advent of computational and numerical methods in recent times has provided new avenues for analyzing art historiographical narratives and tracing the evolution of art styles therein. Here, we investigate an evolutionary process underpinning the emergence and stylization of contemporary user-generated visual art styles using the complexity-entropy (\textit{C}-\textit{H}) plane, which quantifies local structures in paintings. Informatizing 149,780 images curated in DeviantArt and Bēhance platforms from 2010 to 2020, we analyze the relationship between local information of the \textit{C}-\textit{H} space and multi-level image features generated by a deep neural network and a feature extraction algorithm. The results reveal significant statistical relationships between the \textit{C}-\textit{H} information of visual artistic styles and the dissimilarities of the multi-level image features over time within groups of artworks. By disclosing a particular \textit{C}-\textit{H} region where the diversity of image representations is noticeably manifested, our analyses reveal an empirical condition of emerging styles that are both novel in the \textit{C}-\textit{H} plane and characterized by greater stylistic diversity. Our research shows that visual art analyses combined with physics-inspired methodologies and machine learning, can provide macroscopic insights into quantitatively mapping relevant characteristics of an evolutionary process underpinning the creative stylization of uncharted visual arts of given groups and time.
\end{abstract}

\begin{center}
\end{center}
\vspace{0.35cm}

] 




\section{\textbf{Introduction}}
The contemporary visual arts scene embraces a wide array of artistic forms and creative standards that are gaining popularity through information-sharing platforms. While there have been significant historical examinations of how creative visual art styles have evolved, the quantitative analysis of creative stylization is a relatively recent development. We introduce a research design that aims to document the characteristics of the emergence of artistic movements and their subsequent stylizations. By testing more refined empirical data, we improve upon the previous statistical physics approach. Our interdisciplinary research design incorporates two distinct fields.

First, theoretical definitions of creativity have identified two sequential phases in the recognition of creativity. In the first phase, creativity is seen as “the intentional arrangement of cultural and material elements in a way that is unexpected for a given audience \cite{godart2020sociology}.” In the following phase, the creativity finds its “relevance in the new context,” where it contributes to the solution of a problem it assiduously formulated \cite{amabile1983social}. While the first phase that “caused scandal” may not necessarily result in the second phase, it often serves as a catalyst for the development of novel artistic styles. Art movements like Surrealism and Dadaism, among others, have used this disruption to challenge conventional norms and spark creative revolutions \cite{hobsbawm1995age}.

Upon the concatenation of the initial disruption and the emergence of a new style, a transformative process takes place. This process influences the creative acts of subsequent artists through the observation, analysis, and internalization of the new style. By assimilating and building upon these innovations, artists contribute to the dynamic evolution of art \cite{aoki2011rates,enquist2008does}. Therefore, to gain a comprehensive understanding of the creative process, it is essential to consider the two phases separately. By analyzing the initial novelty and the emergence of a new style as distinct phenomena, we can better appreciate the individualized contributions of each artwork as an original mutation. This analytical approach allows for a more nuanced interpretation of artistic development within an enhanced spatiotemporal framework.

Second, while theoretical studies of creativity in the domain of art can be traced back to earlier eras, contemporary efforts to quantitatively assess and measure relevant characteristics have only recently gained momentum \cite{park2020novelty,elgammal2015quantifying}. Through the development of computational and robust methods for analyzing visual arts and the accumulation of large-scale digital art images, scientists have characterized and uncovered various interesting features within a broad range of artworks. Computational studies of visual art have provided both synchronic and diachronic insights into the zeitgeist and evolution of visual arts. Previous studies of visual art styles have been advanced with statistical data of spatial parameters, including chromatic properties \cite{lee2018heterogeneity}, fractals \cite{taylor1999fractal}, wavelets \cite{lyu2004digital}, the compression ensemble approach \cite{karjus2023compression}, and concepts of entropy \cite{kim2014large,sigaki2018history,lee2020dissecting}. 

In particular, complexity (\textit{C})-entropy (\textit{H}) plane as a statistical tool has proven to intuitively map a diachronic evolution of art-historical styles \cite{sigaki2018history,perc2020beauty}. In statistical physics, measurements of \textit{C} and \textit{H} based on the distribution of local ordering patterns were initially used to characterize time-series signals; these measurements have since been generalized to analyze patterns of 2D images \cite{ribeiro2012complexity}. \textit{C} represents the degree to which the local order patterns deviate from both a random and homogeneous distribution, and \textit{H} represents the degree of disorder in the image's pixel organization. \textit{C} reflects the degree to which the objects within an image are spatially bounded or interrelated (e.g., Impressionist paintings tend to have low \textit{C} due to the use of different types of brush patterns; different local patterns are detected uniformly on a painting), whereas \textit{H} reflects the degree to which the objects of an image are more clearly outlined or exhibit fluidity between them (e.g., Mondrian’s minimal geometric paintings tend to have low \textit{H}) (refer to Methods; complexity-entropy (\textit{C}-\textit{H}) measures of visual artwork images).

Sigaki \textit{et al}.’s approach \cite{sigaki2018history} appraised the localization of the \textit{C} and \textit{H} of 137,364 visual artwork images from WikiArt, and gained a view of the hierarchical clustering structure of 92 art-historical styles (between Renaissance and Contemporary/Postmodern Art) in the \textit{C}-\textit{H} plane. The study regarded the dynamics of transitions of different artistic styles in the \textit{C}-\textit{H} plane as evolutionary. This suggests that decoding the latent dynamics that have driven the path of the styles in the \textit{C}-\textit{H} plane would reveal the evolutionary process underpinning the emergence and stylization of a specific group of visual arts in terms of the two aforementioned phases of creativity and stylization. Yet, the deterministic properties of the localized information of artworks in the bounded \textit{C}-\textit{H} space raises a question of how to estimate an evidence of creative stylization of visual arts in the \textit{C}-\textit{H} plane.

Here, we consider the indication of diverse image representations of a specific timeframe and groups (i.e., intragroup image diversity) as a conditional phase of the visual artistic stylization in the \textit{C}-\textit{H} plane. In this context, we conduct an exploratory data analysis (EDA) \cite{sahoo2019exploratory} on stylization of 149,780 visual art images from quasi-canonical “user-generated arts \cite{yazdani2017quantifying}” of DeviantArt and Bēhance platforms of a given timeframe (2010-2020) in the \textit{C}-\textit{H} plane, as an extension of the temporal evolution of art styles by Sigaki \textit{et al}. Through empirically identifying the relationships between the intragroup image diversity and the corresponding local information of the \textit{C}-\textit{H} space, we hypothesize that there will be a particular \textit{C}-\textit{H} movement (i.e., mutations) of a specific group of visual artworks over a given timeframe for their stylistic diversity to manifest in the \textit{C}-\textit{H} plane. To test this hypothesis, we ask the following research questions.

\textit{RQ 1.} Can the \textit{C}-\textit{H} plane, which effectively characterized the styles of art-historical paintings, also capture the temporal stylistic evolution (i.e., \textit{C}-\textit{H} trajectories) of contemporary user-generated visual arts? If so, what would these trajectories represent? 

\textit{RQ 2.} How is the average \textit{C}-\textit{H} position of user-generated visual arts at a given time related to the intragroup image diversity? Can we comprehend the relationship between \textit{C}-\textit{H} positions and the intragroup image diversity through the local information of the \textit{C}-\textit{H} space, specifically focusing on the diversity of image representations that can be expressed in particular regions?

To address the issues in the aforementioned questions, we conduct the following analyses. 

1) Through validating the applicability of our image set conveyed by the \textit{C}-\textit{H} measures and their robustness, we reveal characteristics of temporal stylistic transitions of quasi-canonical user-generated visual arts from DeviantArt and Bēhance in the \textit{C}-\textit{H} plane. Moreover, we explore sub-visual art fields within the user-generated visual arts that have significantly influenced the temporal stylistic transitions in the \textit{C}-\textit{H} plane.

2) We use two types of similarity measures---cosine similarity and Jaccard similarity---on two multiple image representation spaces---image embeddings through a pre-trained Residual Network (ResNet) architecture \cite{yilma2023elements} and Scale-invariant Feature Transform (SIFT) features \cite{lowe2004distinctive}, to measure the image diversity/degree of dissimilarity in a specific \textit{C}-\textit{H} region. The low- (e.g., visual elements such as lines, contours, height, edges, angles, dots, colors, etc.) and high-level image features (e.g., themes of shapes and objects comprised of low-level features) extracted from the two methods aggregately encompass art style-level explicability of images \cite{liu2021understanding}. 

In addition, we use the autoregressive moving-average (ARMA) model to statistically examine a temporal relationship between the average \textit{C}-\textit{H} positions of artworks within a given period and the average dissimilarities of their multi-level image features. Finally, we investigate the diversity of image representations that can be expressed in different areas of the \textit{C}-\textit{H} space and unveil an empirical condition for the emergence of styles that are not only novel in the \textit{C}-\textit{H} plane, but also characterized by greater stylistic diversity.   


\section{\textbf{Results}}
\subsection{\textbf{Stylization of curated visual arts in the \textit{C}-\textit{H} plane: DeviantArt and Bēhance (2010-2020)}}
To account for the stylization of contemporary visual artworks mainly distributed via online communication channels of information, we investigate DeviantArt and Bēhance, serving as massive online visual art platforms involving various creative fields. Since there are huge collections of artworks on both platforms, we specifically focus on representative and quasi-canonical subsets of daily promoted user-generated visual artworks (“\textit{Daily Deviation}” and “\textit{Best of Bēhance}”) curated by those platforms. This allowed us to process manageable and significant amounts of data \cite{akdag2013flow,buter2011explorative}.  The image set used for complexity-entropy (\textit{C}-\textit{H}) measurement is described in Table SI \ref{table S1.}.

We map the yearly average \textit{C} and \textit{H} values of 149,780 user-generated visual artwork images from DeviantArt and Bēhance (2010 - 2020) (see Methods; Complexity-entropy (\textit{C}-\textit{H}) measures of visual artwork images, for a detailed explanation of the calculation of \textit{C} and \textit{H} values). For comparison, we also overlay the \textit{C}-\textit{H} values of conventional art historical paintings from the WikiArt dataset \cite{tan2018improved} as Sigaki \textit{et al}.’s experiment \cite{sigaki2018history}. The WikiArt dataset is composed of 26,415 paintings, primarily collected from Western art history, spanning the period of 1301 to 2016 CE. The art historical periods depicted in our projection of the evolution of visual arts are virtually identical to those identified by Sigaki \textit{et al}.  

\begin{figure*}[!ht]
\centering
\includegraphics[width=16cm]{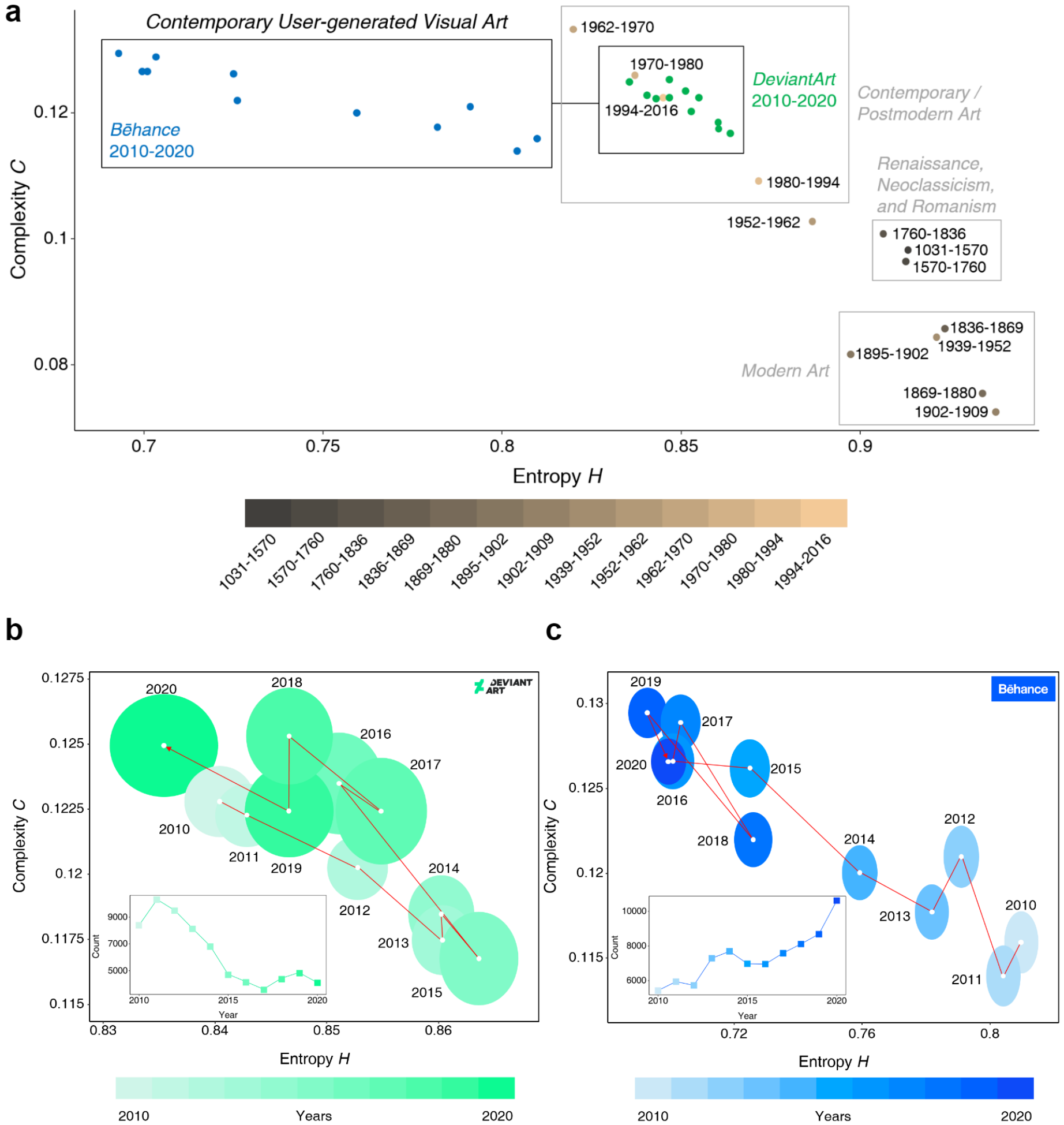}
\caption{\textbf{Group-level temporal transitions of visual art styles in the complexity-entropy (\textit{C}-\textit{H}) plane.} (\textbf{a}) Contemporary user-generated visual art styles of DeviantArt and Bēhance (2010 - 2020) compared to the main divisions of art historical periods mapped in the \textit{C}-\textit{H} plane. 
Images with distinct forms and a limited number of ordinal patterns typically exhibit low \textit{H} values and high \textit{C} values. Conversely, disordered images comprising more random or interrelated components generally display high \textit{H} values and low \textit{C} values. The result reveals that the average \textit{C}-\textit{H} values of art historical styles have been shifted towards the higher \textit{C} and lower \textit{H} areas over time. (\textbf{b}, \textbf{c}) Yearly \textit{C}-\textit{H} trajectory of contemporary user-generated visual arts from DeviantArt’s \textit{Daily Deviations} and Bēhance’s the \textit{Best of Bēhance} (refer to Data (Materials)). Each \textit{C}-\textit{H} plane is composed of multiple elliptical areas, with each ellipse representing 95\% confidence interval bounds from the yearly average \textit{C}-\textit{H} values of the intragroup visual artworks. The multiple CI ellipses and the yearly \textit{C}-\textit{H} trajectories connecting them altogether describe yearly transitions of visual art styles within each platform. Also, insets are included to show the cumulative number of samples over the given time frame.}
\label{Fig. 1 a-c}
\end{figure*}

Fig. \ref{Fig. 1 a-c} demonstrates that both platforms’ \textit{C}-\textit{H} trajectories tend towards the upper-left \textit{C}-\textit{H} region compared to the previous periods, although their average position shows a slight difference (refer to Table SI \ref{table S2.} for more information regarding the difference in the overall positions of the two platforms in the \textit{C}-\textit{H} plane).

Also, a closer look at the \textit{C}-\textit{H} values of DeviantArt and Bēhance reveals that the averaged positions of the two platforms have varied significantly over time. We cross-validated the temporal difference in \textit{C}-\textit{H} values on each platform by evaluating the predictive accuracies of our dataset’s \textit{C}-\textit{H} values using four machine learning algorithms. All the classification models predict the curation year of paintings with probabilities (DeviantArt: $\sim$15\%, Bēhance: $\sim$13.7\%) significantly higher than the chance level (Fig. SI \ref{Fig. S1}). 

The \textit{C}-\textit{H} trajectories of temporal transitions of DeviantArt and Bēhance visual art styles over a decade extend that of the transition between early 20th century and 1970s (e.g., from painterly/optic to linear/haptic). Observing the beginning and ending years (Fig. \ref{Fig. 1 a-c}b, c), the \textit{C} of DeviantArt’s visual art increased by 0.002 (from 0.123 to 0.125), while their \textit{H} decreased by 0.005 (from 0.84 to 0.835). In the case of Bēhance, there was a rise in \textit{C} by 0.01 (from 0.12 to 0.13) and a decline in \textit{H} by 0.11 (from 0.81 to 0.7) over the same period. The group-level temporal transitions of visual art styles among the two platforms reveal indirect but eventual tendencies towards styles with higher \textit{C} and lower \textit{H}. Consequently, the chronological \textit{C}-\textit{H} trajectories of the contemporary user-generated visual arts of both platforms from 2010 to 2020 indicate that their stylizations altered in a macroscopically similar manner, despite the differences in direction details within each platform. 

Meanwhile, the yearly average \textit{C}-\textit{H} values of Bēhance are notably distinct from those of DeviantArt and the major art historical periods. We explore which sub-visual art fields substantially influenced the \textit{C}-\textit{H} positions of Bēhance over time by separately examining the temporal \textit{C}-\textit{H} movements of five creative fields that account for the greatest proportion of samples in the Bēhance dataset. We observe the yearly average \textit{C}-\textit{H} values of samples from Illustration, Graphic Design, Character Design, Animation, and Fine Arts fields in Bēhance. Fig. SI \ref{Fig. S2 a.b} shows that all five fields’ yearly average \textit{C}-\textit{H} values are specifically directed towards a similar \textit{C}-\textit{H} area (\textit{C}: 0.1 - 0.3, \textit{H}: 0.65 - 0.85) through a decade of movements. Overall, the \textit{C}-\textit{H} movements of the fields reveal a tendency for homogeneous transitions of visual art styles in Bēhance during the time period. 

Extending Sigaki \textit{et al}.’s analysis on the evolution of art styles allows us to discover the group-level temporal stylizations of quasi-canonical visual artworks in DeviantArt and Bēhance (2010 – 2020) aligning with that of the modern art identified by Sigaki \textit{et al}. Overall, the \textit{C}-\textit{H} movements of the platforms are characterized by an increase in the high-level \cite{ullman2000high} visual structure of artworks incorporating geometrical object-oriented patterns; the emergence and taking up of an artistic style with clearer and simpler visual elements. 

Through structural notions of creativity and stylization, it is possible to infer the strengthened stylization of the contemporary user-generated visual arts towards a more linear/haptic region in the \textit{C}-\textit{H} plane. This trend can be interpreted as a process where artworks exhibit what Birkhoff called a “bijectively homomorphic” (structurally similar) \cite{birkhoff1940lattice} stylization. In simpler terms, while the visual representations of artworks remain diverse and random at an individual level, they interact and gradually align with one another, leading to a homogenization of styles. The significance of individual artists' innovation being proportional to the extent of its influence on others \cite{galenson2010understanding}, fuels a competitive and mimetic process under uncertainty, paradoxically resulting in a cycle of imitation as a necessary step toward creativity \cite{de1903laws}. Also, in any given institution, the number of “insiders” on the curatorial committee is fewer than expected, resulting in “a durable and recognizable pattern of aesthetic choices \cite{godart2018style}”. In response to the inherent “symbolic uncertainty” in the platform of visual arts, individual artists may be subject to “mimetic isomorphism,” or mimicking others’ distinguished artistry as the most cost-effective strategy for gaining insider acceptance \cite{dimaggio1983iron}. 

The convergence of two distinct groups of images within the \textit{C}-\textit{H} framework over a decade exemplifies mimetic isomorphism, where shared environmental factors (e.g., technological trends, aesthetics, or cultural shifts) drive contextual adaptations rather than mere replication. This phenomenon highlights how stylistic and structural similarities can emerge across distinct entities through imitative and adaptive processes in complex creative domains. The rapid dissemination of information and images, driven by platform-based visual content structures, might have accelerated the homogenization of contemporary user-generated visual art styles.


\subsection{\textbf{Spatiotemporal relationships between image diversity and the \textit{C}-\textit{H} space}}

We investigate how the average \textit{C}-\textit{H} positions of user-generated visual arts from DeviantArt and Bēhance moved over time. In this section, we attempt to reveal the spatiotemporal relationships between image diversity and the local information of the \textit{C}-\textit{H} space, so that we can understand the previous \textit{C}-\textit{H} movements over time in terms of intragroup diversity and stylization. 

\begin{figure*}[!ht]
\centering
\includegraphics[width=15cm]{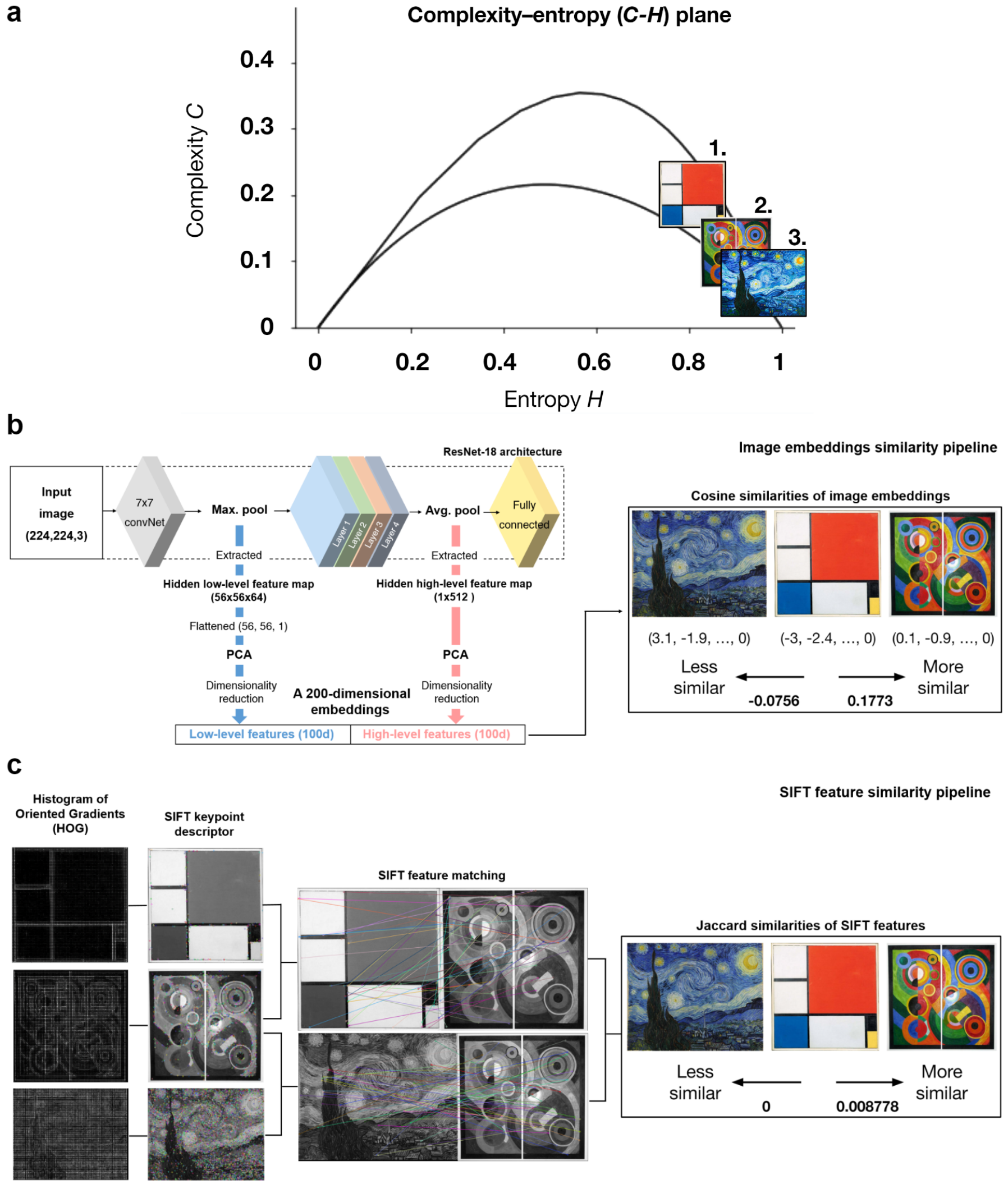}
\caption{\textbf{Image representations in the complexity-entropy (\textit{C}-\textit{H}) plane and pipelines of image similarity measures through different image representation spaces.} (\textbf{a}) Three reputable exemplary masterpieces chosen and localized in the \textit{C}-\textit{H} plane according to the given embedding parameters ($d_x = d_y = 2$), to aid viewers’ understandings of the image representations the plane encompasses. (\textbf{b}) A schematic diagram of an image similarity computation pipeline through a neural network. We measure pairwise cosine similarities of a 200-dimensional embedding vector including both low- and high-level feature maps extracted from individual visual artworks. (\textbf{c}) A schematic diagram of the image similarity computation pipeline through the SIFT descriptor. We measure pairwise Jaccard similarities of SIFT descriptor matching a 128-dimensional vector of low-level features extracted from individual visual artworks. In addition to the image representations the \textit{C}-\textit{H} plane encompasses, we adopt and use two different types of image processing methods to obtain multi-level image features and measure their similarities: ResNet architecture and SIFT algorithm. Through our approach, we investigate characteristics of image representations in the \textit{C}-\textit{H} plane with image similarity measures that aggregate both low- and high-level features. The implemented images (1 – 3.) are “Composition 2” by Piet Mondrian, 1929 (Public Domain, retrieved from WikiArt, https://www.wikiart.org/en/piet-mondrian/composition-2), “Rhythm” by Robert Delaunay, 1912 (Public Domain, retrieved from WikiArt, https://www.wikiart.org/en/robert-delaunay/rhythm-1), and “The Starry Night” by Vincent van Gogh, 1889 (Public Domain, retrieved from WikiArt, https://www.wikiart.org/en/vincent-van-gogh/the-starry-night-1889). All the used image samples are available in the public domain.}
\label{Fig. 2 a-c}
\end{figure*}

In order to assess the intragroup diversity of image representations from the two platforms in the \textit{C}-\textit{H} space, we use two different similarity measures on two types of image features. First, we use cosine similarity of the artwork’s image embeddings (IE): multi-level image representations comprising both low- (100-dimensional) and high-level (100-dimensional) image features. Further, we use the Jaccard similarity coefficient of low-level features (i.e., keypoints) of two images matched by the standard SIFT algorithm. Fig. \ref{Fig. 2 a-c}  shows the detailed pipelines of the two image similarity measures. The analytical independence of the two similarity measures and the distance in the \textit{C}-\textit{H} plane verifies that their pairwise correlations are fairly weak (range: 0.003 - 0.07; see Fig. SI \ref{Fig. S3 a-f}).


We first examine the relationship between the average \textit{C}-\textit{H} positions of user-generated visual arts from DeviantArt and Bēhance over the specified timeframe (2010-2020) and their intragroup image diversity. To account for trend components in the time-series data, we use an ARMA model to regress the average values from the dyadic similarities among the artworks on the average values of the complexity \textit{C} and entropy \textit{H} over the given years. 


\begin{figure*}[!ht]
\centering
\includegraphics[width=17cm]{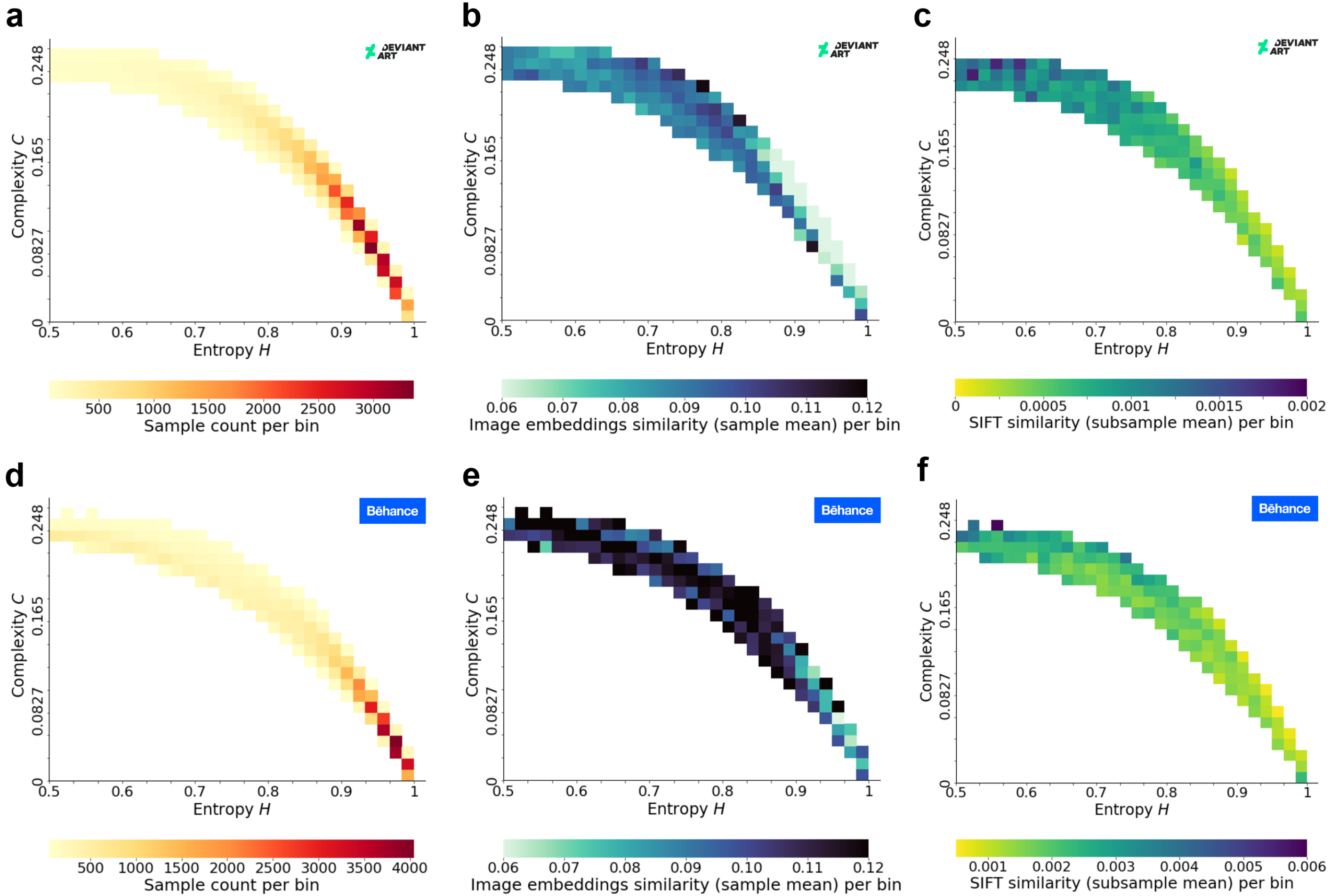}
\caption{\textbf{Visualization of spatial relationships between the \textit{C}-\textit{H} space and the observed image diversity in DeviantArt and Bēhance data as measured by the image similarity.} (\textbf{a}, \textbf{b}, \textbf{c}) Degrees of sample density, IE similarity (sample mean), and SIFT similarity (subsample mean) of the DeviantArt images per bin in the \textit{C}-\textit{H} region. (\textbf{d}, \textbf{e}, \textbf{f}) Degrees of sample density, IE similarity (sample mean), and SIFT similarity (subsample mean) of the Bēhance images per bin in the \textit{C}-\textit{H} region. Each \textit{C}-\textit{H} bin (\textit{C} $\approx$ 0.01, \textit{H} $\approx$ 0.02) per platform occupies more than 50 images.}
\label{Fig. 3 a-f}
\end{figure*}


The ARMA model is applied separately to subsampled images from the DeviantArt and Bēhance datasets. The results (Table \ref{table 1.}) indicate that the \textit{C} and \textit{H} values of a particular year are associated with a decrease in similarity, suggesting that the \textit{C} and \textit{H} values can affect the increase in diversity, despite the decrease in average absolute diversity over time. Here, we note that as the analysis predicts a movement of the average value of diversity for a given year with its average value of \textit{C} and \textit{H}, the area with large \textit{C} and \textit{H} values in the entire \textit{C}-\textit{H} space does not represent the area of high diversity. Meanwhile, the variance of \textit{H} is added as an independent variable and controlled on the one hand, confirming that \textit{H} variance also signiﬁcantly has a positive correlation with diversity in all cases.



\begin{table} [H]
  \centering
  \renewcommand{\arraystretch}{1.5}
  \resizebox{\columnwidth}{!}{
  \begin{tabular}{lllll}
    \multicolumn{5}{l}{\textbf{Regressions Predicting Similarity with ARMA Errors (2010-2020)}}\\ 
    \cmidrule(r){1-5}
    \textbf{Variables}     & \textbf{IE Similarity}  &\textbf{IE Similarity}   &\textbf{SIFT Similarity}   &\textbf{SIFT Similarity} \\
                      & \textbf{(DeviantArt)}  &\textbf{(Bēhance)}   &\textbf{(DeviantArt)}   &\textbf{(Bēhance)} \\
                      &
                      \multicolumn{2}{c}{($p = 3$, $q = 1$)} &
                      \multicolumn{2}{c}{($p = 1$, $q = 1$)} \\
    \midrule 
    \\
    Entropy \textit{H} Mean & -3.650***  & -1.018***  & -0.043***  &-0.135*  \\
    					& (0.250)  & (0.177)  & (0.004)  & (0.055)  \\ [0.5cm]
    Entropy \textit{H} Variance & -5.842***  & -1.632***  & -0.087***  &-0.278**  \\
    					    & (0.770)      & (0.376)       & (0.010)      & (0.105)  \\ [0.5cm]
   Complexity \textit{C} Mean & -4.934***  & -3.801***  & -0.061***  &-0.263+  \\
    					    & (0.342)      & (0.400)      & (0.006)      & (0.149)  \\ [0.5cm]
Complexity \textit{C} Variance& -34.101***& 3.834**     & 0.288*       &0.476  \\
    					    & (5.930)      & (1.423)      & (0.115)      & (0.687)  \\ [0.5cm]
                              Constant & 4.012***    & 1.396***    & 0.045***    & 0.150*  \\
    					    & (0.258)      & (0.200)      & (0.005)      & (0.067)  \\ [0.5cm]
                          Obervations & 11             & 11             & 11             &11  \\
    \\
    \bottomrule 
   \multicolumn{5}{l}{Semirobust Standard errors in parentheses: *** p<0.001, ** p<0.01, * p<0.05 * + p<0.1} 
  \end{tabular}
 }
 \caption{\textbf{ARMA predicting average similarity among visual arts in DeviantArt and Bēhance. Refer to Methods; Statistical regression of visual artwork image similarities for detailed description.}}
\label{table 1.}
\end{table}

Higher \textit{H} is associated with lesser degrees of IE and SIFT similarity among the artworks over time, as revealed by both platforms of our research. Similarly, higher levels of \textit{C} are significantly associated with lesser degrees of IE and SIFT similarity. The \textit{H} of artworks in both platforms are skewed to the left (DeviantArt: skewness = -1.95; Bēhance: skewness = -1.04), whereas the \textit{C} is relatively symmetrical (DeviantArt: skewness = 0.2; Bēhance: skewness = 0.02). 

A distribution of given years’ images in the \textit{C}-\textit{H} space is likely to include a wider variety of styles along with an increase in image diversity within the given year group upon the following conditions. First, when an average movement (direction of increases in \textit{C} and \textit{H}) occurs toward the upper-right \textit{C}-\textit{H} region, which is the optimal direction of the upper-left (similar, unexplored, sparse) and lower-right (diverse, explored, dense) \textit{C}-\textit{H} region. Additionally, when the distribution (large \textit{H} variance) is made on the entropy \textit{H} axis.

We now partially examine the robustness of the ARMA results through patterns of the local stylistic diversity visualized over binned areas of a partitioned \textit{C}-\textit{H} region (\textit{C}: 0 - 0.31, \textit{H}: 0.5 - 1; see Fig. \ref{Fig. 3 a-f}). Overall, the sample count plots (Fig. \ref{Fig. 3 a-f}a, d) show that the most densely populated area is to the lower-right \textit{C}-\textit{H} region with given embedding parameters, $d_x = d_y = 2$. This confirms that there is likely to be a strong correlation between image diversity and the highly concentrated \textit{C}-\textit{H} region, where previous art-historical styles with great image diversity are positioned. Hence, the more artworks where entropy \textit{H} is greater, the greater image diversity there is likely to be. 

In terms of observing image diversity manifested in the \textit{C}-\textit{H} region, we first measure the mean value of pairwise cosine similarity of IE of all the samples from each accountable \textit{C}-\textit{H} bin. Subsequently, we measure the mean value of pairwise Jaccard similarity of SIFT features of randomly subsampled artworks from each accountable \textit{C}-\textit{H} bin. The results altogether reveal that the highly dense \textit{C}-\textit{H} area tends to reflect less image similarity. The IE similarity (Fig. \ref{Fig. 3 a-f}b, e) and SIFT similarity (Fig. \ref{Fig. 3 a-f}c, f) plots demonstrate that at a given level of entropy \textit{H}, higher complexity \textit{C} is likely to be associated with greater intragroup image diversity (also see Fig. SI \ref{Fig. S4 a-f} for the supporting results of the WikiArt images previously mapped for Fig. \ref{Fig. 1 a-c}a). 

As observed in Fig. \ref{Fig. 1 a-c}a, our results uncover the \textit{C}-\textit{H} movements of quasi-canonical visual artworks in contemporary online platforms, which ultimately trended to the upper-left \textit{C}-\textit{H} region. The \textit{C}-\textit{H} movements begin and propel away from the highly dense and stylistically diverse (low \textit{C} and high \textit{H}) and move towards sparser and stylistically homogenic areas (high \textit{C} and low \textit{H}), forming a process of a particular stylization. Such a stylization (towards higher \textit{C} and lower \textit{H}) is consistent with the broader movements in Western art history \cite{lee2020dissecting}, which ranged from classical to modern times. However, the region with high \textit{C} and low \textit{H} has the lowest diversity (i.e., high image similarity) of image representations in artworks. Therefore, even though propelling towards the area is a new artistic endeavor in terms of \textit{C} and \textit{H}, there are few artworks that could be expressed with the given \textit{C}-\textit{H} values.

Based on our analyses, we suggest conditions for the groups of visual artworks to have high intragroup diversity and shed light on an evolutionary process in which the intragroup diversity gives rise to styles. It is confirmed that when the average \textit{C}-\textit{H} movement of the groups is balanced towards higher \textit{C} and higher \textit{H} rather than the extremes in the \textit{C}-\textit{H} space, the stylistic diversity of the groups shows a significant tendency to increase more compared to the previous period. Our findings also suggest that a visual artistic stylization process occurs as a result of the cumulative diversity of individualized artworks contributing as original mutations over time, to the transformative process of the prior narratives of styles and the emergence of a new style.


\section{\textbf{Discussion}}
Beyond computationally observing the macroscopic evolution of visual art, opportunities for analytically mapping its latent dynamics in a creativity framework remain. Notably, Sigaki \textit{et al}. \cite{sigaki2018history} have performed a seminal work that demonstrates the feasibility of using the \textit{C}-\textit{H} plane to intuitively map the diachronic evolution of visual art styles. The study has recognized “a natural law in the same way as physical growth (Wölfflin)” when affiliating historical alterations in style, particularly from linear to painterly. This remark is similar to David Bohm’s belief about artistic creativity being related to “a harmony parallel to that of nature (Cézanne) \cite{bohm2004creativity}.” 

This study quantitatively tracks user-generated artworks of the contemporary visual art disciplines within the \textit{C}-\textit{H} space and interprets their stylization from the perspective of diversity within groups. Apart from the findings of Sigaki \textit{et al}.’s study providing useful insights into the evolution of visual art styles and their potential relationship to natural laws and creativity, we take an alternative strategy to explore the evolutionary dynamics of creative visual arts in the \textit{C}-\textit{H} plane. We look at the evolutionary process underpinning a stylization through the lens of selection, in which a variety of competing agents interact. This state arises from the emergence of novel and different mutations during the reproduction process \cite{nowak2006evolutionary}. A stylization of visual arts entails the development of popularization of diverse artistic forms influenced by novel stylistic canons.  Therefore, considering diversity as a prerequisite for the emergence of creativity and subsequent stylization, we investigate the \textit{C}-\textit{H} plane to empirically identify the optimal \textit{C}-\textit{H} condition, by which stylistic diversity is most pronounced while visual artworks of a specific timeframe and groups continue migrating towards homogenic styles.  

We find that as the average entropy \textit{H} rises, so does the intragroup diversity. Our analyses indicate that a novel style emerges in \textit{C}-\textit{H} regions where random and diverse agents coexist with strong individual innovativeness (Fig. \ref{Fig. 1 a-c}). This finding echoes Eric Hobsbawm, who observed that during the period of highly unpredictable social and technological changes (i.e., the Avant Garde prior to 1914), modern art (low \textit{C} and high \textit{H}; Fig. \ref{Fig. 1 a-c}a) was “not to claim that it displaced the classic and the fashionable, but that it supplemented both \cite{hobsbawm1995age}.” 

The ARMA analysis (Table \ref{table 1.}) supports our viewpoint by demonstrating a substantial increase in the respective group’s intragroup diversity when the group's \textit{C}-\textit{H} movement is directed towards a balance (synchronous increases in \textit{C} and \textit{H}). Moreover, the results of this study empirically confirm that the intragroup image diversity varies depending on the locations in the \textit{C}-\textit{H} plane, further indicating the heterogeneity of image representations and local stylistic diversity of the \textit{C}-\textit{H} space. Therefore, from a collective rather than an individualist perspective, we assume that the optimal \textit{C}-\textit{H} condition for the greatest intragroup diversity and emergence of a certain style will be revealed when a group sets a balance between attempts to escape the conventionally dense \textit{C}-\textit{H} area (moving toward high \textit{C} and low \textit{H} areas) and attempts to remain in the \textit{C}-\textit{H} area where diversity can be expressed (low \textit{C} and high \textit{H}). 

Building on the study by Sigaki \textit{et al}. \cite{sigaki2018history}, our research provides a novel perspective on contemporary visual art disciplines by positioning user-generated artworks within the complexity-entropy (\textit{C}-\textit{H}) space. This framework quantifies stylization while revealing diversity and coherence within artistic groups. Our work extends the analysis of \textit{C}-\textit{H} trajectories of visual art stylization from the early 20th century to the 1970s, confirming an intensified tendency of contemporary visual artworks to shift toward the upper-left region within the given \textit{C}-\textit{H} framework. By situating our findings within art history, parallels can be drawn with earlier work that used neural networks to trace artistic influences across historical movements \cite{narag2021discovering}, as both approaches highlight how stylistic diversity reflects creative innovation and adherence to group norms. Similarly, our entropy-based analysis aligns with a study that applied spatial entropy to landscape analysis \cite{papadimitriou2022spatial}, underscoring entropy's utility in analyzing structural and compositional diversity across different contexts. 

Our research also complements the findings of Valensise \textit{et al}. \cite{valensise2021entropy}, who used the \textit{C}-\textit{H} plane to analyze meme evolution, highlighting how community-driven curation fosters a balance between novelty and conformity. This dynamic is evident in curated subsets like DeviantArt’s “\textit{Daily Deviation}” and Bēhance’s “\textit{Best of Bēhance},” which mirror the evolutionary patterns of digital culture. Additionally, our work aligns with Deng \textit{et al}. \cite{deng2020exploring}, who examined representativity in art paintings. Their study underscores the challenges of quantifying art’s diversity while preserving its nuanced cultural significance. By extending these efforts to contemporary digital art, we bridge historical and modern perspectives, demonstrating how user-generated art mirrors broader trends in collective artistic evolution. 

Meanwhile, our study has several limitations, including potential selection bias in our data and the methodologies used for measuring image similarity. While our dataset includes works by artists from around the world, it is predominantly composed of two-dimensional visual artworks created by Western artists. Furthermore, the curated user-generated artworks on DeviantArt and Bēhance during the specified time period likely represent only a fraction of the vast and dynamic evolution occurring within contemporary visual art disciplines.

As platforms showcasing global creativity, their curated collections are expected to result in cultural and social diversity. However, many independently operated platforms within specific cultural contexts remain unexplored, presenting valuable opportunities for further research. Future studies could address these limitations by incorporating a broader range of data sources, including other social media and regional art-sharing platforms, to enhance the representation of global artistic and cultural diversity, and to validate and expand upon our findings.

Regarding our image feature extraction methods, despite the fact that the ResNet-18 architecture and SIFT capture multi-level image features, there are numerous other possible combinations of methods that capture various spatial image characteristics. Understanding the limitations of our image similarity measurement methods and how they impact our analyses is crucial. The SIFT algorithm is sensitive to image variations (e.g., significant affine transformations, illumination changes, and image noise) possibly affecting the detection and description of keypoints. Also, possible loss of subtle yet important features or fine-grained details cannot be ignored due to reduced dimensionality in ResNet latent PCA embeddings. Therefore, exploring alternative or complementary visual features in future experiments could provide more robust and generalizable insights.

The rapid advancements in computer vision have introduced a variety of image processing models, each offering unique capabilities for analyzing visual data. Different choices of feature extraction models may produce moderately varying similarity outcomes, thereby shaping diversity assessments and influencing our understanding of the evolution of art styles. Notably, transformer-based models, such as Vision Transformers (ViTs) \cite{dosovitskiy2020image}, have demonstrated potential for capturing global context across entire images, providing richer and more nuanced feature embeddings. Specifically, CLIP (Contrastive Language-Image Pretaining) \cite{radford2021learning} and DINO (Distillation of Non-contrastive Image Representations) \cite{caron2021emerging} provide embeddings aligned with semantic meaning, which could improve high-level similarity detection. While our research focused on assessing the “diversity of visual forms” by focusing more on visual information itself, future studies could explore the diversity and evolution of contemporary visual arts from a new perspective. This could involve utilizing state-of-the-art models to measure artwork similarity based on semantic and contextual information, offering a deeper understanding of artistic trends and styles.

Future research could also investigate the properties of images generated by artificial intelligence (AI) models. In recent years, state-of-the-art AI models (e.g., DALL-E2 \cite{ramesh2022hierarchical}, stable diffusion models \cite{rombach2022high}, etc.) have made enormous strides in text-to-image art generation. With growing interest in AI-generated images, the relationship between AI and human creativity has come to the forefront of research. A recent study of images generated by a stable diffusion model revealed that AI-generated images tend to have a narrower distribution of entropy and complexity when compared to human drawings; this may lead to less diversity and creativity in their visual artistic expression \cite{papia2023entropy}. In addition, another study addressed the difficulties of evaluating interactions between human and computational creativity in online creative ecosystems \cite{oppenlaender2022creativity}. Even though the development of AI art is rapidly advancing and attracting a great deal of attention, we cannot overlook the need to investigate and comprehend its characteristics in a more systematic manner. One could pose queries about the potentials of AI art, with a desire to investigate its representations and inspirations. As our methods are readily applicable to any corpus of two-dimensional digital images and their resulting knowledge maps, we believe the framework of this study can facilitate a better understanding of diversity and stylization in the emerging field of AI art.


\section{\textbf{Data (materials)}}
Quantitatively analyzing raw cultural data of user-generated content to capture their similarities on many possible dimensions can yield insights into the diversity of data’s visual organization in multidimensional spaces \cite{manovich2020cultural}. On this account, we assess popular online platforms as sources of artistic creativity and diversity - DeviantArt and Bēhance, which have become increasingly popular among artists who exhibit and share their creative works online. Accessibility, availability, and direct observation of a large corpus of user-generated visual art images are notable benefits of these platforms. 

 Meanwhile, the vast size of the entire collection of user-generated artworks on these platforms makes comprehensive analysis impractical. Therefore, we processed a more manageable and meaningful dataset of high-quality artworks broadly recognized as influential within their respective artistic communities. To ensure that the images analyzed represented broad trends in contemporary user-generated visual art, we specifically focused on subsets of artworks curated by the platforms themselves: “\textit{Daily Deviation}” on DeviantArt and “\textit{Best of Bēhance}” on Bēhance. These subsets are quasi-canonical, as they are promoted daily by the platforms based on quality, innovation, and relevance, reflecting current trends in user-generated art.

We use the 149,780 images with the \textit{C}-\textit{H} values (Table SI \ref{table S1.}) during the IE obtaining process. However, the computing process of extraction and pairwise comparison of SIFT features between images consumes considerably more time compared to that of the image embeddings through convolutional neural networks (CNN). Therefore, we take two separate subsampling procedures for EDA on spatiotemporal relationships between local information of the \textit{C}-\textit{H} space and image diversity. 1) We randomly subsample images for both IE and SIFT similarity measures for the ARMA model (Table SI \ref{table S3.}). 2) Considering each \textit{C}-\textit{H} bin sample as a population, we additionally use the sample size formula (confidence level: 95\%, margin of error: 5\%) to determine and randomly subsample the minimum number of necessary samples to meet the desired statistical constraints, and to draw proper inferences from respective \textit{C}-\textit{H} bins in Fig. \ref{Fig. 3 a-f}.


\section{\textbf{Methods}}
\subsection{\textbf{Complexity-entropy (\textit{C}-\textit{H}) measures of visual artwork images}}
Sigaki \textit{et al}. \cite{sigaki2018history} observed that changes in complexity (\textit{C}) and entropy (\textit{H}) of paintings over time could reflect the evolution of art historical styles. Here, we explain how a two-dimensional image is represented in the \textit{C}-\textit{H} space. We calculate the normalized permutation entropy \textit{H} and statistical complexity \textit{C} of an image based on its matrix representation. The original image files are in JPEG and PNG formats, represented in a 24-bit RGB color space (8 bits each for the red, green, and blue channels). We obtain a simpler matrix representation of each image by averaging the three color channels of every pixel.

Next, we examine submatrices with ordinal patterns for each image using embedding dimensions $d_x \times d_y$ (in our case, $d_x = d_y = 2$), leading to ($(d_x d_y)! = 24$) possible ordinal patterns. By sliding partitions of size $d_x \times d_y$ pixels across the entire matrix, we obtain a distribution of ordinal patterns $P$ for the image. Here, $P=\{p_i  ;i=1,\dots,n\}$ can be viewed as a 24-dimensional vector whose elements sum to 1.

From the probability distribution $P$, we calculate the normalized Shannon entropy $H(P)$:

\begin{equation}\label{1}
H(P)=\frac{1}{\ln (n)} \sum_{i=1}^n p_i \ln \left(1 / p_i\right)
\end{equation}

, where $n = (d_x d_y)!$.
        
The entropy \textit{H} approaches 1 if the pixel order appears random, and $H \approx 0$ if the pixel order is highly regular and always appears in the same configuration. The value of \textit{H} signifies the degree of ‘disorder’ in the configuration of the pixels in an image, as represented by its matrix representation. 

Furthermore, $C(P)$ is calculated to investigate the degree of structural complexity present in the submatrices. $C(P)$ is defined as:

\begin{equation}\label{2}
C(P)=\frac{D(P, U) H(P)}{D^*}
\end{equation}

, where $U = \{ u_i = 1/n \; ; \; i = 1, \dots, n \}$ represents the uniform distribution, $D (P, U)$ is the Jensen-Shannon divergence between $P$ and $U$, and $D^*$ is the maximum Jensen-Shannon divergence.

The Jensen-Shannon divergence $D (P, U)$ is given by:

\begin{equation}\label{3}
D(P, U) = S\left( \frac{P + U}{2} \right) - \frac{S(P)}{2} - \frac{S(U)}{2}
\end{equation}

, where $(P + U)/2 = \left\{(p_i + 1/n){2},\; i = 1, \dots, n \right\}$. The maximum divergence $D^*$ can be calculated as: 

\begin{equation}\label{4}
D^* = \max D(P, U) = \frac{1}{2} \left[ \frac{n+1}{n} \ln(n+1) + \ln(n) - 2 \ln(2n) \right]
\end{equation}

This is obtained when $P = \{ p_i = \delta_{1,i} \; ; \; i = 1, \dots, n \}$, where $\delta_{1,i}$ is the Kronecker delta.

$C(P)$ increases as the distribution $P$ of the local order patterns deviates from the uniform distribution $U$ or as the entropy of $P$ increases. Specifically, $C(P)$ reaches its maximum in a state that is neither completely uniform nor entirely homogeneous. Stylistically, $C(P)$ reflects how much the objects within an image are spatially circumscribed or interrelated, while $H(P)$ reflects how distinctly the objects are outlined or how fluidly they are intertwined \cite{perc2020beauty}. 

In practice, we calculate the \textit{C}-\textit{H} values of artworks using the Ordpy module \cite{pessa2021ordpy}, a simple and open-source Python module that implements permutation entropy and several principle methods for analyzing time series and two-dimensional data using complexity parameters \cite{ribeiro2012complexity,hao2020gaze,osmane2019jensen}. To obtain the upper- and lower-boundary curves in the \textit{C}-\textit{H} plane for our 2D images, we employed two functions \texttt{maximum\_complexity\_entropy} and \texttt{minimum\_complexity\_entropy} in the Ordpy module \cite{pessa2021ordpy}. These functions systematically generate multiple probability distributions over the possible ordinal patterns and compute both $H(P)$ and $C(P)$. They do so by assigning probability mass to one or a small number of ordinal patterns while distributing the remaining probability evenly among the other patterns, scanning through this space of distributions for \textit{H} and \textit{C} values. Collecting and sorting the \textit{H} and \textit{C} points yields approximate upper- and lower-boundary curves that envelop the values found in experimental or simulated data, thus delineating the limits in the \textit{C}-\textit{H} plane.

Lastly, for the choice of embedding dimensions, we set $d_x=d_y=2$. While these values are tuning parameters that can be adjusted, it is known from previous studies that $(d_x d_y )!$ $\ll$ (width × height) must hold to ensure reliable results \cite{ribeiro2012complexity}. Given that the average image width and height of 149,780 images used in this study are approximately 1,000 pixels (average width: 1011.61px, average height: 953.91px; refer to Table SI \ref{table S1.}), we chose $d_x=d_y=2$ as the embedding dimensions. 


\subsection{\textbf{Statistical regression of visual artwork image similarities}}

Among typical statistical models for time series analysis (e.g., the Bayesian network, the hidden Markov model, etc.), we used the ARMA model to regress the dyadic similarities among the artworks on the degree of complexity \textit{C}, entropy \textit{H}, and their variances over the respective years from 2010 to 2020 (Table \ref{table 1.}):

\begin{equation}\label{5}
y_t = X_t \beta + \sum_{i=1}^{p} \phi_i y_{t-i} + \sum_{i=1}^{q} \theta_i \varepsilon_{t-i} + \varepsilon_t
\end{equation}

, where $y_t$ is the average value of the dyadic similarities in DeviantArt and Bēhance at year $t$, and $X_t$ is a vector of the \textit{C}-\textit{H} covariates at $t$. $p$ and $q$ are the number of lags for autocorrelation and moving-average, respectively. In order to obtain a stationary series of our data, we used the modified Dickey-Fuller unit-root test \cite{cheung1995practitioners}, which determined the parameter $p$. This results in $p = 3$ for IE models, and $p = 1$ for SIFT models. $q$ was determined to be 1 for both IE and SIFT models using Bartlett’s approximation \cite{box2015time}.


\subsection{\textbf{Multi-level image features for visual artwork similarity measures}}
The Euclidean distance in the \textit{C}-\textit{H} space (Fig. \ref{Fig. 2 a-c}a) is a straightforward metric for measuring the stylistic dissimilarity between images. In addition, we further adopt and utilize both global and local descriptors to extract multi-level image features and measure their similarity: a ResNet architecture and the SIFT algorithm. By utilizing the multi-level image features that aggregate both low- and high-level features, we disclose stylistic characteristics of image representations in the \textit{C}-\textit{H} plane. 

Recent analyses of visual art styles use style vectors based on deep CNN to extract stylistic (i.e., high-level) features and information from paintings \cite{gatys2015neural}. On the other hand, the SIFT feature descriptor as a well-established CV technique has been demonstrated that the descriptor is effective for a variety of objectives related to image matching and object recognition. CNN filters by themselves perform similarly to SIFT in detecting low-level and straightforward (i.e., handcrafted) local invariant features, while outperforming SIFT in detecting high-level and complex features \cite{zheng2017sift}. The combined usage of CNN and SIFT features has proven to result in discriminative multi-level image features using the best of both \cite{yan2016cnn}.

We initially build on ResNet-18 architecture pretrained with the ImageNet database to construct multi-level image features in our image set. ResNet pretrained with ImageNet has also been shown to provide sufficient feature representations for paintings \cite{yilma2023elements}, making it appropriate for extracting latent visual features - i.e. image embeddings (IE)---from heterogeneous visual artworks.   

Inspired by Liu \textit{et al}.’s approach \cite{liu2021understanding} of extracting multi-level features of artworks to represent art styles of images, we obtain the embeddings of our images in the following manner (see Fig. \ref{Fig. 2 a-c}b for the detailed technical pipeline). 1) As combining both low- and high-level features is effective for encoding art styles, the max (front) and average (back) pooling layers of ResNet-18 are chosen as the convolutional layers to extract hidden feature maps from each input image (224 × 224 × 3). 2) We then reduce the dimensionality of the hidden low-level feature map (56 × 56 × 64) extracted from the maximum pooling layer to a one-dimensional feature map (56 × 56 × 1). 3) We also extract a hidden high-level feature map (1 × 512) from the average pooling layer just prior to the fully connected layer, where no feature map exists. 4) We then run a Principal Component Analysis (PCA) on both low-level and high-level feature maps to derive 100-dimensional embedding from each, thereby creating a 200-dimensional embedding vector of each user-generated artwork. 5) Lastly, we measure pairwise cosine similarities between embeddings of the contemporary user-generated visual artworks. 

Following the neural network-based image similarity measurement, we utilize the SIFT feature matching technique (see Fig. \ref{Fig. 2 a-c}c for the detailed technical pipeline). Specifically, SIFT uses the local histogram of oriented gradients to match the local features of an image with those of other images \cite{szeliski2022computer}. The SIFT features are extracted from keypoints detected between two distinct images while preserving their original data dimensions; gradient orientation histograms are extracted from quadrants of points of interest in an image. They are then merged into a normalized histogram (a SIFT descriptor) that is invariant to image location, scale, and rotation \cite{lowe2004distinctive}. Each SIFT descriptor in an image is compared to its counterparts in other images.

We implement the standard SIFT feature indexing and matching technique for obtaining the desired degree of similarity between images based on its highest matching accuracy compared to the performance of other succeeding feature descriptors (e.g., PCA-SIFT, SURF, BRIEF, ORB, etc.) \cite{karami2017image}. As the SIFT feature extractor, we use the xfeatures2d module from the cv2 interface in OpenCV versions for Python \cite{citekey} to calculate the SIFT matching degree between images. 1) We first detect the local extrema in a single image as a potential keypoint detected in each pixel compared with its neighboring pixels. Each accurate keypoint selected from an image then generates an orientation histogram of 8 bins for the keypoints’ neighboring 16 × 16 blocks, which are then subdivided into 16 sub-blocks of size 4 × 4. The respective keypoints are subsequently allocated 128 (= 8 × 16) bin values as a vector \cite{sri2022detecting}. 2) We then calculate the number of acceptable matches between keypoints from a pair of input images by the Fast Library for Approximate Nearest Neighbors (FLANN) based on a k-nearest neighbors (KNN) matcher \cite{vijayan2019flann}. 3) Using the number of keypoints and their good matches from each pair of images, we calculate the intersection (\textit{n} of matching keypoints) divided by the union (\textit{n} of total keypoints from a pair of images) of the images’ keypoints.  


\section*{\textbf{Data availability}}
The data underlying the analyses and findings of this study will be available from S.K. (ryankim1101@kaist.ac.kr) on a reasonable request. Researchers interested in accessing data can contact the corresponding author (wnjlee@kaist.ac.kr). Correspondence and requests for data should be addressed to S.K. and W.L.

\section*{\textbf{Competing interests}}
The authors declare no competing interests.


\normalsize
\bibliography{main}


\section*{\textbf{Acknowledgements}}
We would like to thank Lev Manovich, Kangsan Lee, and Frédéric Godart for their valuable feedback and constructive suggestions on earlier versions of this manuscript. We sincerely thank the anonymous reviewers for their helpful comments to improve this manuscript. S.K., B.L., and W.L. acknowledge the support from the BK21 Plus Postgraduate Organization for Content Science (N20210014). 


\onecolumn

\newpage
\vspace*{1cm}
\begin{center}
    \LARGE \textbf{Supplementary Information} \\
    \vspace{0.5cm}
    \large \textbf{Figures and tables}
\end{center}

\vspace{6cm} 

\setcounter{table}{0}
\renewcommand{\tablename}{Table SI}
\begin{table}[H]
  \centering
  \resizebox{\columnwidth}{!}{
  \begin{tabular}{lllllllll}
   \toprule
\textbf{Platform} & \textbf{Data type} & \textbf{Initial}      & \textbf{Measurement} & \textbf{Uncomputable}  & \textbf{Zero \textit{C}-\textit{H}} & \textbf{Processed} & \textbf{Filtered}   & \textbf{Image} \\
                        &                           & \textbf{samples}   &                                  & \textbf{samples}           & \textbf{samples}   & \textbf{samples}   &  \textbf{samples}  & \textbf{dimensions}  \\
  \midrule

         DeviantArt &                           & 70,804     &                                              & 210                             & 4                        & 70,590                & 68,735                 & Width (avg.): 1013.15px,\\ 
                        & JPEG                   &               & Complexity-entropy                   &                                   &                           &                         &                           & Height (avg.): 1023.82px \\                     
                        & and                     &               & (\textit{C}-\textit{H}) values       &                                   &                           &                         &                           &                                  \\ 
                        & PNG                    &               & per image                               &                                   &                           &                         &                           & Width (avg.): 1010.31px,\\ 
         Bēhance   &                           & 348,040    &                                              & 9,939                           & 2,685                  & 335,416              &  81,045                & Height (avg.): 894.61px  \\
  \bottomrule
           Total.     &                            & 418,844 &                                                & 10,149                         & 2,689                   & 406,006 ($\approx$ 97\%)                 & 149,780 ($\approx$ 36\%)       & $\approx$ 1,000px  \\         
  \bottomrule
\end{tabular}
}
 \caption{\textbf{Dataset description for complexity-entropy (\textit{C}-\textit{H}) measure.} We directly downloaded user-generated visual artwork images via image URLs extracted from the web pages of the corresponding projects that have been curated and promoted on a daily basis between early 2010 and the end of 2020 in DeviantArt (\textit{Daily Deviations}) and Bēhance (\textit{Best of Bēhance}). Following the extraction of JPEG and PNG images from our initial image set for image quantification, our initial sample consists of a total of 418,844 artworks, including 70,804 images from DeviantArt and 348,040 images from Bēhance. Subsequently, we generate annual time series data for empirical studies in this paper by combining all the image data values into yearly sample datasets based on corresponding platforms and the years of the original projects’ showcase dates. As for the \textit{C}-\textit{H} measures, we exclude 10,149 images that could not be computed by the cv2 module as ‘abnormal’ sets, as well as 2,689 images with both \textit{C}-\textit{H} values of 0. The rest of the 406,006 images with respective \textit{C} and \textit{H} values are then filtered to 149,780 images of sub-visual art fields of DeviantArt and Bēhance. Given that the average image width and height of 149,780 images used in this study are approximately 1,000 pixels (average width: 1011.61px, average height: 953.91px), we chose $d_x=d_y=2$ as the embedding dimensions.}
\label{table S1.}
\end{table}


\clearpage

\setcounter{figure}{0}
\renewcommand{\figurename}{Figure SI}
\begin{figure*}
\centering
\includegraphics[width=18cm]{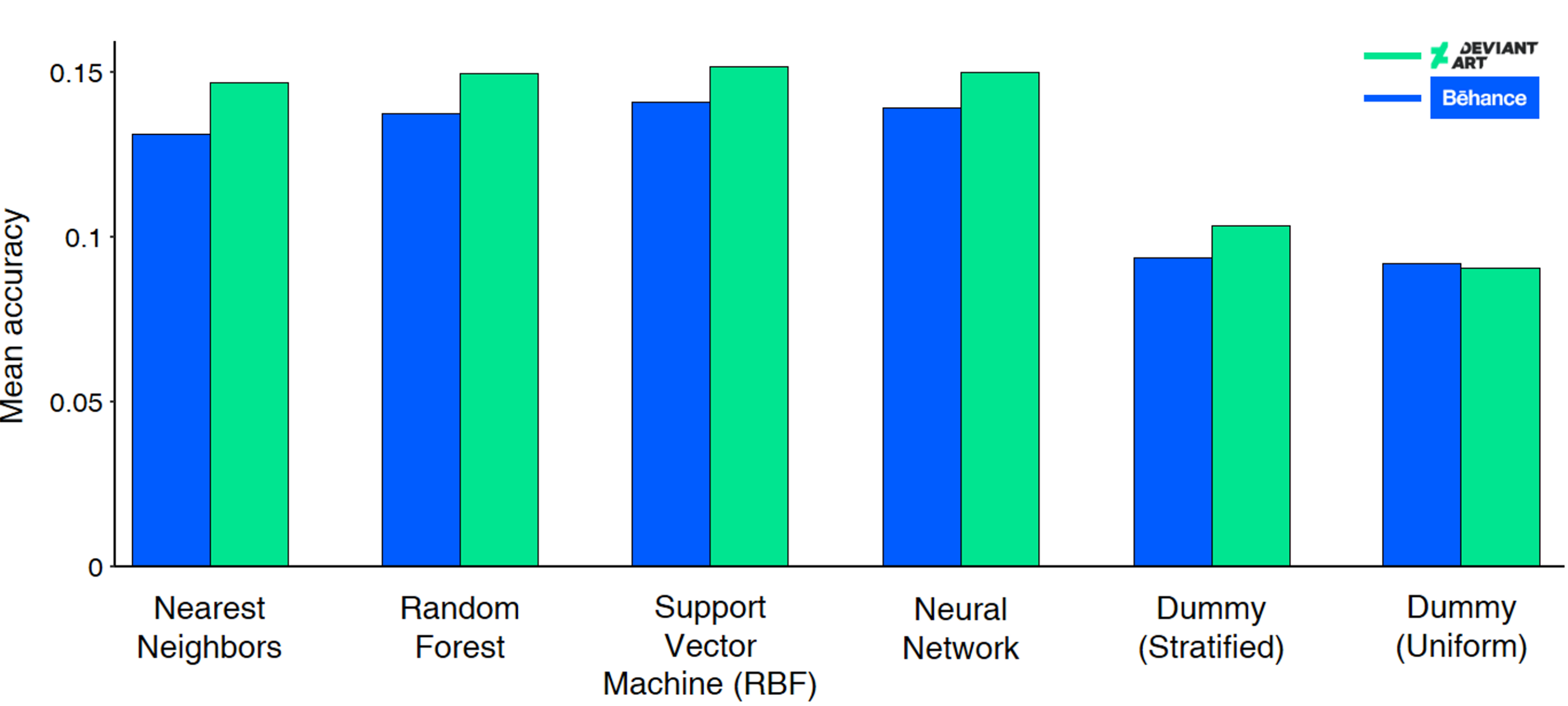}
\caption{\textbf{Cross validating predictive accuracies of statistical learning algorithms.} For a robustness check, we cross-validate the predictive accuracy of \textit{C}-\textit{H} values from our dataset using four machine learning algorithms (nearest neighbors, random forest, support vector machine, and neural network). We consider the classification of \textit{C}-\textit{H}  values of diverse artworks based on their original platforms and the year of curation. As a result, it is confirmed that the accuracy scores of the overall classifiers are considerably higher than those of the dummy classifiers: stratified $\approx$ 10\% (DeviantArt), $\approx$ 9.3 - 9.4\% (Bēhance), and uniform $\approx$ 9.1 - 9.2\% (DeviantArt), 9.3 - 9.4\% (Bēhance)). Each platform's statistical algorithms have comparable mean accuracy (DeviantArt: $\sim$ 15\% and Bēhance$\sim$ 13.7\%). In line with cross-validation results regarding \textit{C}-\textit{H} measures from Sigaki \textit{et al}., our results demonstrate that the \textit{C}-\textit{H} measures of contemporary visual artworks accurately predict both the years and platforms to which they belong. This implies a descriptive evaluation of the degree to which individual images correspond to each year label in the \textit{C}-\textit{H} plane.}
\label{Fig. S1}
\end{figure*}
\FloatBarrier

\clearpage

\begin{table}
\centering
\renewcommand*{\arraystretch}{1.5}
\begin{tabular}{l|l}
\textbf{Creative fields - Bēhance}   & \textbf{N of samples (prop.)} \\ \hline
3D Art                               & 263 (0.32\%)                    \\
Architecture                         & 5,562 (6.86\%)                  \\
Crafts                               & 1,735 (2.14\%)                  \\
Fashion                              & 4,728 (5.83\%)                  \\
Fine Arts                            & 6,611 (8.16\%)                  \\
Game Design                          & 1,355 (1.67\%)                  \\
Graphic Design                       & 27,591 (34.04\%)                \\
Illustration                         & 59,048 (72.86\%)                \\
Photography                          & 4,434 (5.47\%)                  \\
UI/UX / Information Architecture     & 1,708 (2.11\%)                  \\
Animation                            & 6,043 (7.46\%)                  \\
Cartooning                           & 3,091 (3.81\%)                  \\
Character Design                     & 19,459 (24.01\%)               
\end{tabular}
\caption{\textbf{Number of samples associated with clustered sub-topics of Bēhance---normalized proportion of samples in each creative field over the total number of samples in Bēhance with each sample associated with multiple creative fields.} The difference in the overall positions of the two platforms in the \textit{C}-\textit{H} plane is due to the different proportions of genres that make up each platform. The majority of the Bēhance image set consists of images from Graphic Design and Illustration, while the DeviantArt image set is mostly composed of paintings featuring pictorial objects. Table SI \ref{table S1.} summarizes the distribution of creative fields in the Bēhance dataset. In the case of DeviantArt, due to the absence of image classification from the individual project pages in the platform, the categorization of images is manually conducted by examining a sample size of 1,000 images. Consequently, 72.1\% of the images are drawings or paintings containing pictorial objects, and the proportion of photos and design works is notably small.}
\label{table S2.}
\end{table}
\FloatBarrier

\clearpage

\begin{figure*}
\centering
\includegraphics[width=18cm]{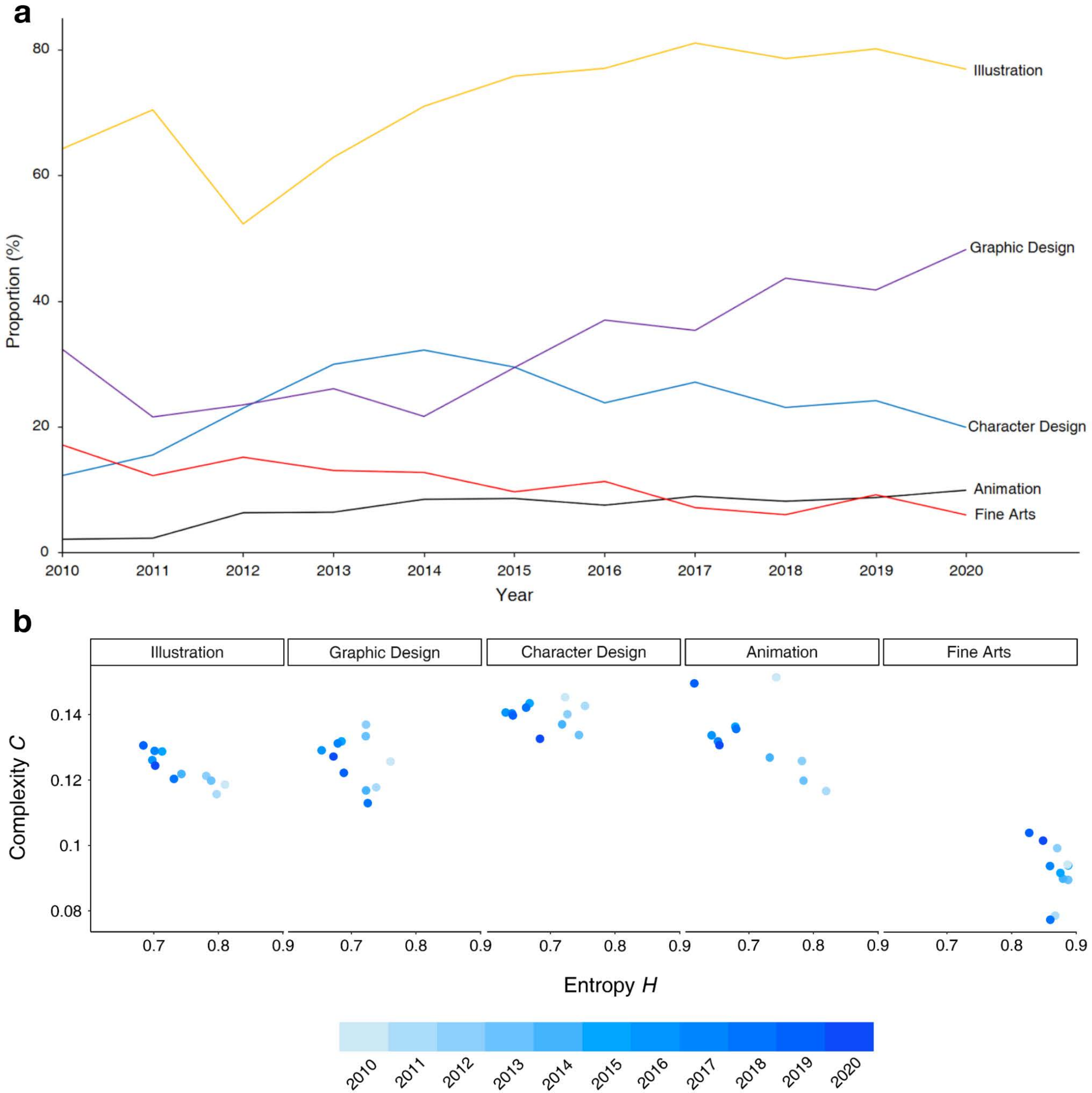}
\caption{\textbf{Five creative fields of Bēhance with the greatest sample proportion and their annual mean \textit{C}-\textit{H} values (2010-2020).} (\textbf{a}) Yearly trends of sample proportions from the creative fields (Illustration, Graphic Design, Character Design, Animation, and Fine Arts) in Bēhance (2010-2020). Comparing samples from 2010 and 2020, the annual proportion of the Illustration field in 2010 (64.27\%, \textit{n} = 3,491) increase by 4,680 images, accounting for 76.92\% (\textit{n} = 8,171) of samples from 2020. The Graphic Design field’s annual proportion in 2010 (32.36\%, \textit{n} = 1,758) increased by 3,368 images, making it the second highest proportion of the year 2020 samples (48.25\%, \textit{n} = 5,126). The other two categories are Character Design and Animation, which increase by 1,451 images and 939 images, respectively, making them the third (19.95\%, \textit{n} = 2,119) and fourth (9.5\%, \textit{n} = 1,057) proportional fields by 2020. In 2010, the annual percentage and sample size for the Fine Arts field (12.79\%, \textit{n} = 984) decreased to 6.06\% (\textit{n} = 644). (\textbf{b}) Yearly average \textit{C}-\textit{H} values of visual artworks from the creative fields (Illustration, Graphic Design, Character Design, Animation, and Fine Arts) in Bēhance (2010-2020). Concerning the five fields’ annual \textit{C}-\textit{H} movements, the annual average \textit{H} values of their samples decrease over a decade, indicating that their styles become more linear. Both the annual average \textit{C} and \textit{H} values for Animation and Character Design decrease over time, indicating that they become less complex and more linear. In addition, the Illustration, Graphic Design, and Fine Arts sectors exhibit an increase in their annual average \textit{C} values over time, indicating that their styles incorporate incremental continuums of visually interconnected elements.}
\label{Fig. S2 a.b}
\end{figure*}
\FloatBarrier

\clearpage

\begin{figure*}
\centering
\includegraphics[width=18cm]{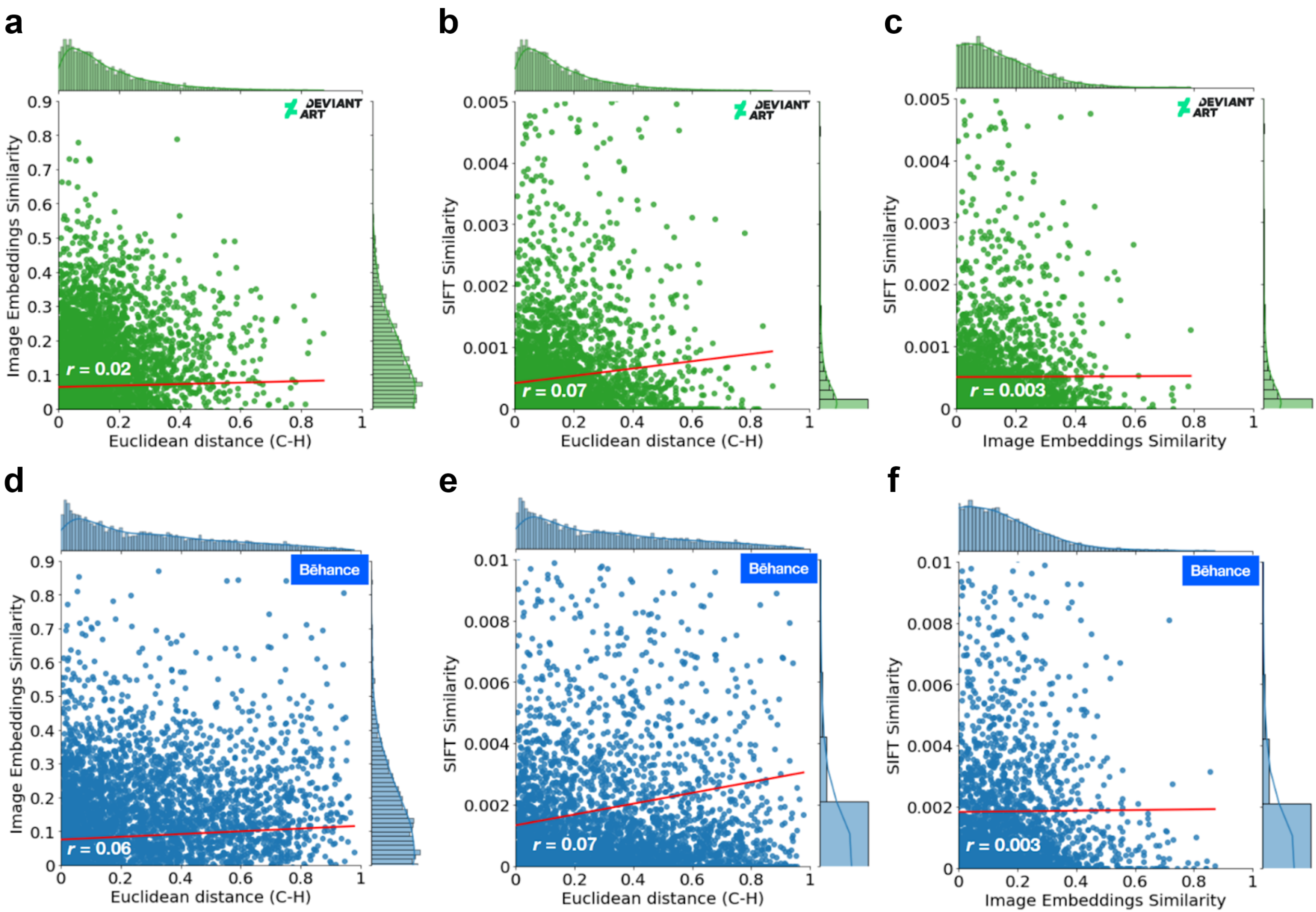}
\caption{\textbf{Pairwise correlations between image similarity measures.} (\textbf{a}, \textbf{b}, \textbf{c}) Pairwise correlations between IE similarity, SIFT similarity, and Euclidean distance (\textit{C}-\textit{H}) of image representations in 5,000 pairs of DeviantArt artworks. (\textbf{d}, \textbf{e}, \textbf{f}) Pairwise correlations between IE similarity, SIFT similarity, and Euclidean distance (\textit{C}-\textit{H}) of image representations in 5,000 pairs of Bēhance artworks. We verify correlations between image similarity measures used in this paper. Marginal plots with regression lines of two different similarity measures of 5,000 unique pairs of visual art images across the entire \textit{C}-\textit{H} plane (\textit{C}: 0 - 1, \textit{H}: 0 - 1) altogether show weak correlations (refer to correlation coefficients in each plot) in DeviantArt and Bēhance.}
\label{Fig. S3 a-f}
\end{figure*}
\FloatBarrier

\clearpage

\begin{table}
\centering
\renewcommand*{\arraystretch}{1.5}
\begin{tabular}{lll}
\textbf{Measurement} &
  \textbf{N of random subsamples} &
  \textbf{Processed pairs} \\ \hline
\begin{tabular}[c]{@{}l@{}}Pairwise cosine similarities of \\ IE of images\end{tabular} &
  \begin{tabular}[c]{@{}l@{}}11,000 images per platform\\ (1,000 images per year)\end{tabular} &
  \begin{tabular}[c]{@{}l@{}}5,494,500 pairs per platform \\ (499,500 pairs per year)\end{tabular} \\ \hline
\begin{tabular}[c]{@{}l@{}}Pairwise Jaccard similarities of \\ SIFT features of images\end{tabular} &
  \begin{tabular}[c]{@{}l@{}}1,100 images per platform\\ (100 images per year)\end{tabular} &
  \begin{tabular}[c]{@{}l@{}}19,800 pairs per platform\\ (4,950 pairs per year)\end{tabular} \\ \hline
\end{tabular}
\caption{\textbf{Dataset description of image similarity measures used for the ARMA model.} To calculate the average image similarity from the online platforms for each given year, we randomly subsample 1) 1,000 images from each year image set and calculate the mean of the 499,500 pairwise IE similarities, and 2) 100 images from each year image set and calculate the mean of the 4,950 pairwise SIFT similarities.}
\label{table S3.}
\end{table}
\FloatBarrier

\begin{figure*}
\centering
\includegraphics[width=18cm]{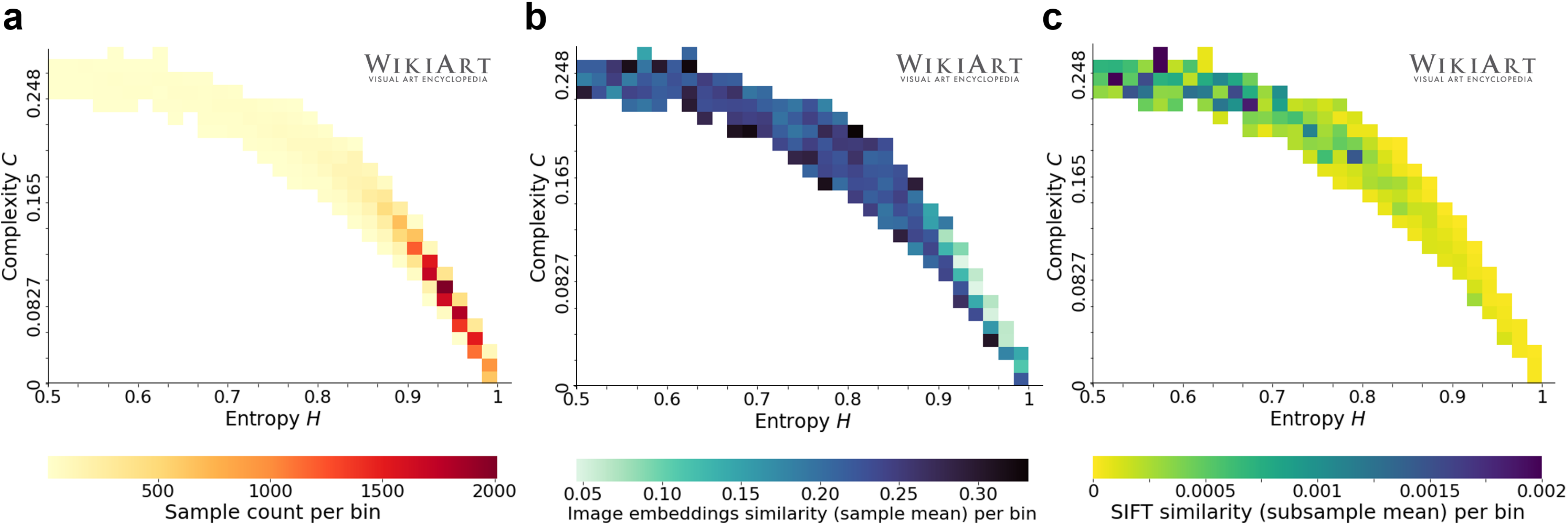}
\caption{\textbf{Visualization of spatial relationships between the \textit{C}-\textit{H} space and the observed image diversity in WikiArt data as measured by the image dissimilarity.} (\textbf{a}, \textbf{b}, \textbf{c}) Degrees of sample density, IE similarity (sample mean), and SIFT similarity (subsample mean) of the WikiArt images per bin in the \textit{C}-\textit{H} region. Each \textit{C}-\textit{H} bin (\textit{C} $\approx$ 0.01, \textit{H} $\approx$ 0.02) within a partitioned \textit{C}-\textit{H} region (\textit{C}: 0 - 0.31, \textit{H}: 0.5 - 1) per platform occupies more than 10 images.}
\label{Fig. S4 a-f}
\end{figure*}
\FloatBarrier

\FloatBarrier

\end{document}